\documentclass[journal,twoside,web]{IEEEtran}
\usepackage{cite}
\usepackage{amsmath,amssymb,amsfonts}
\usepackage{graphicx}
\usepackage{textcomp}
\usepackage{cite}
\usepackage{booktabs}
\usepackage{multirow}
\usepackage{makecell}
\newcommand{\smallgrey}[1]{{\scalebox{0.8}{#1}}}
\usepackage[linesnumbered,ruled]{algorithm2e}
\usepackage{url}

\begin{document}

\title{A Gradient-based Causal Discovery Framework with Applications to Complex Industrial Processes}

\author{Meiliang Liu, \IEEEmembership{Member, IEEE}, Huiwen Dong, Xiaoxiao Yang, Yunfang Xu, Zijin Li, Zhengye Si, Xinyue Yang, Zhiwen Zhao
\thanks{This work has been granted by National Natural Science Foundation of China under Grant 61075075. \textit{(Corresponding author: Zhiwen Zhao.)}}
\thanks{Meiliang Liu, Huiwen Dong, Xiaoxiao Yang, Yunfang Xu, Zijin Li, Zhengye Si, and Xinyue Yang are with the School of Artificial Intelligence, Beijing Normal University, Beijing, China (e-mail: liumeiliang520@gmail.com; donghw.dhw@gmail.com; yangxiaoxiao@mail.bnu.edu.cn; xuyunfang@mail.bnu.edu.cn; 2291832685@qq.com; sizhengye0302@gmail.com; 202321081046@mail.bnu.edu.cn).}
\thanks{Zhiwen Zhao is with the School of Artificial Intelligence, Beijing Normal University, Beijing, China, also with the Advanced Institute of Natural Sciences, Beijing Normal University, Zhuhai, Guangdong, China (e-mail:mlt.bnu2017@bnu.edu.cn).}
}

\maketitle

\begin{abstract}
With the advancement of deep learning technologies, numerous neural network-based causal discovery models have been proposed. These models can uncover causal relationships among variables and are widely used in modern industrial systems. Although these methods achieve notable improvements, several limitations remain. Most existing approaches adopt the component-wise architecture, requiring the construction of a separate model for each variable, resulting in substantial computational costs. Moreover, imposing sparsity constraints on the first-layer weights of neural networks to infer causal relationships limits their ability to capture complex and nonlinear interactions among variables. To address these challenges, we propose a novel lightweight causal discovery framework, termed Gradient-based Causal Discovery (GCD). Different from conventional component-wise models, GCD only employs a single multilayer perceptron for time series prediction and leverages $\ell_1$ regularization on neural network's input-output gradient to infer causal relationships. Numerical simulations on Lorenz-96, DREAM4, and CausalTime datasets demonstrate that GCD outperforms existing baselines. Furthermore, experiments on three real-world industrial processes, including Tennessee-Eastman, Ultra-processed Food, and Debutanizer, further validate the effectiveness and computational efficiency of GCD, highlighting its applicability to complex industrial processes.
\end{abstract}


\section{Introduction}

\IEEEPARstart{W}{ith} the rapid advancement of industrial technologies, modern industrial processes generate increasingly large-scale, complex, and multi-source data \cite{4-zhang2024regioselective}. Such data complexity poses challenges for inferring intrinsic interactions among variables. In this context, causal discovery, as a powerful tool for uncovering the underlying causal structures among variables, has been extensively applied in industrial analytics, including fault diagnosis, process optimization, and safety assurance \cite{3-wang2023causal}.

Granger causality is a statistical framework for time series causal discovery. It examines whether the historical values of one variable can statistically enhance the prediction of another variable's future values \cite{27granger1969investigating}. Different from correlation-based methods, which capture synchronous activity, Granger causality can identify the temporal order and directional dependence between variables. This distinctive capability enables its application across various complex systems, including industrial processes \cite{1-hua2025integrated,6-sui2025attribution}, climate systems \cite{5-runge2023causal}, and effective connectivity modeling in neuroimaging \cite{liu2025spatiotemporal}.

The advent of deep learning has led to the development of various neural network-based Granger causality models. Tank et al. \cite{1tank2021neural} proposed the component-wise architecture that models each variable with an independent subnetwork, thereby enabling inference of nonlinear causal relationships within multivariate time series. Meanwhile, they also developed two models, component-wise Multilayer Perceptron (cMLP) and component-wise Long Short-Term Memory (cLSTM), which extract the first-layer weights of the MLP/LSTM and apply a sparsity-inducing regularization penalty for causal discovery. Building upon this component-wise architecture, a series of causal discovery frameworks are proposed, including economy statistical recurrent units (eSRU) \cite{3khanna2019economy}, Causal Discovery from Irregular Time Series (CUTS) \cite{9cheng2023cuts}, and Jacobian Granger causality (JGC) \cite{10suryadi2023granger}, etc.

Although these models have achieved some advancements, they still encounter some limitations when applied to complex industrial systems: (1) Most approaches adopt the component-wise architecture, requiring the construction of an independent model for each variable (time series). This design substantially increases computational costs, particularly in industrial systems with tens or hundreds of sensors \cite{1tank2021neural}. (2) Causal relationships are often inferred from the first-layer weights of neural networks. However, this strategy is insufficient for high-dimensional time series because the search space for causal variance becomes broader and may converge to local optima. (3) The intricate network architectures with numerous parameters hinder the scalability and practical deployment of these models in large-scale industrial scenarios.

Motivated by these challenges, we aim to propose a causal discovery framework that eliminates the need for the component-wise architecture by using a single model. Our second objective is to introduce a novel method for inferring causal relationships, rather than relying on the first-layer weights of the neural network. Lastly, we plan to design a simple architecture that ensures scalability and flexibility.

In this paper, we propose Gradient-based Causal Discovery (GCD), a lightweight and efficient framework for time series causal discovery. Our contributions are as follows:
 
\begin{itemize}
\item We introduce a novel causal discovery framework, GCD, which leverages a single MLP and applies $\ell_1$ regularization on the neural network’s input-output gradients to infer causal relationships.
\item Numerical simulations demonstrate that GCD outperforms the existing baselines on the Lorenz-96, DREAM4, and CausalTime datasets.
\item Experiments conducted on the Tennessee-Eastman, Ultra-processed Food, and Debutanizer processes further validate the effectiveness of GCD in uncovering causal structure while achieving reductions in computational cost.
\end{itemize}

\begin{figure*}[h]
\centering
\includegraphics[width=0.95\textwidth]{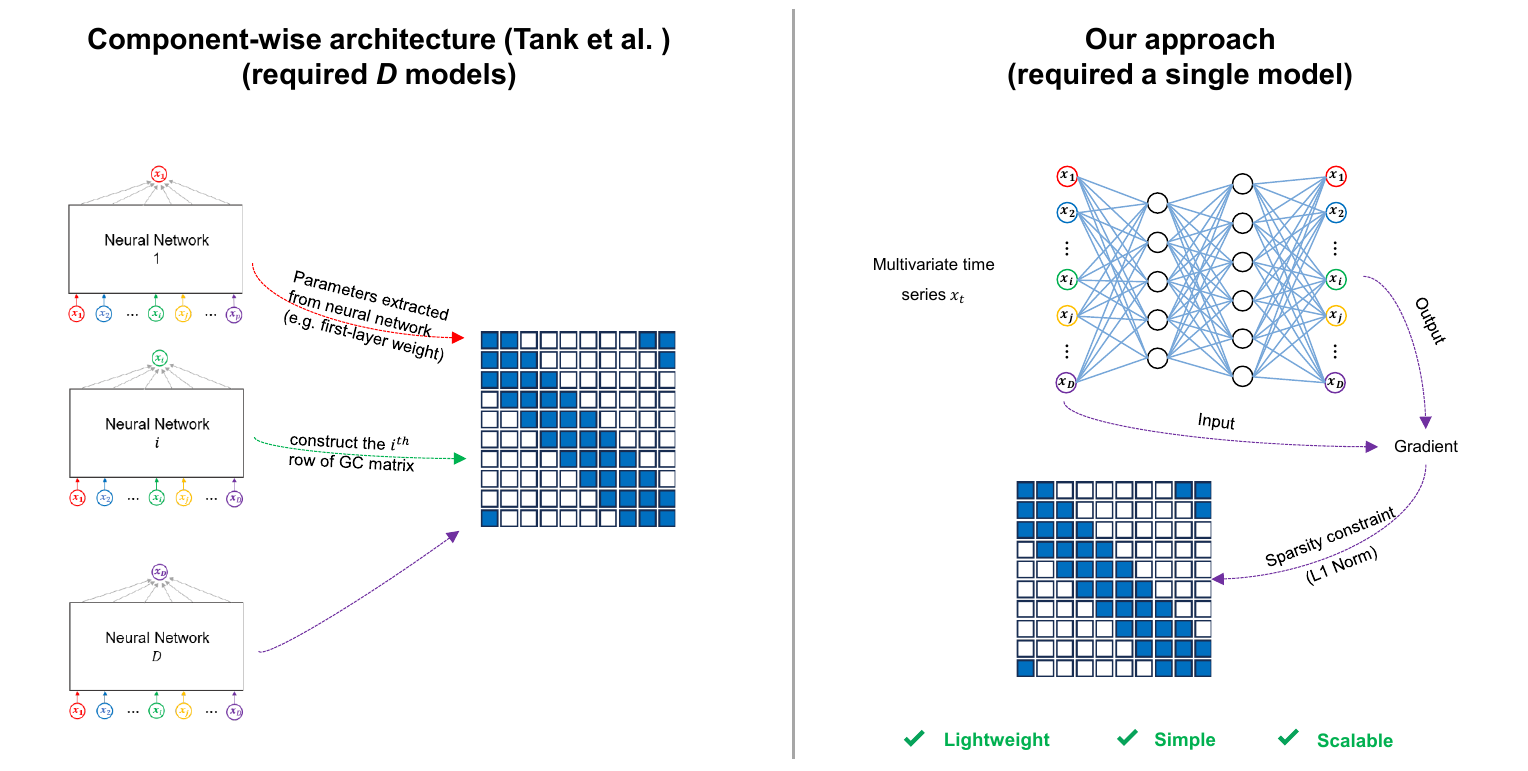} % Reduce the figure size so that it is slightly narrower than the column.
\caption{(Left) The component-wise architecture proposed by Tank et al. \cite{1tank2021neural}, which required $D$ models for $D$-dimensional time series. (Right) The  architecture of proposed GCD, which only required one model for $D$-dimensional time series.}
\end{figure*}

\section{Preliminary}

\subsection{Time series Granger causality inference}

We inherit the notation in Cheng et al. \cite{5cheng2024cuts+} and denote a multivariate time series as $\{x_{t, i}\}_{i=1}^D$, where $t\in\{1,\ldots,T\}$, $i \in\{1,\ldots,D\}$, with $T$ is the total sampling points, $D$ is the dimension of the time series. According to the nonlinear autoregressive model, $x_{t,i}$ can be denoted as a combination of the time series past values:
\begin{equation}
    x_{t,i}=f_{i}(x_{t-k:t-1,1},x_{t-k:t-1,2},\ldots ,x_{t-k:t-1,D})+e_{t,i} 
\end{equation}
where $x_{t-k:t-1,i}$ is the past value of series $i$, $k$ is the maximum time lag, $e_{t,i}$ is an independent noise item, $f_{i}$ usually takes the form of the neural network or other nonlinear function. 

Since the prediction of $x_{t,i}$ depends on the historical information of other time series, if the past values of a series $x_{j}$ can significantly enhance the prediction of the future values of $x_{i}$, then $x_j$ is Granger-cause to $x_i$ ($x_{j}\to x_{i}$); otherwise, $x_j$ is consider no Granger-cause to $x_i$ ($x_{j}\nrightarrow x_{i}$).

\subsection{Previous works}
Neural network-based approaches for discovering causal relationships from nonlinear time series have gained significant attention. Tank et al. \cite{1tank2021neural} proposed cMLP and cLSTM, which utilized the first-layer weights of MLP and LSTM, combined with sparsity-inducing regularization, to discover Granger causality. Khanna et al. \cite{3khanna2019economy} proposed eSRU, applying the sparsity constraints on the first-layer weights of statistical recurrent units (SRU) to infer causal relationships. Bussmann et al. \cite{8bussmann2021neural} proposed Neural Additive Vector Autoregression (NAVAR), including NAVAR(MLP) and NAVAR(LSTM), for causal discovery. Nauta et al. \cite{2nauta2019causal} proposed the Temporal Causal Discovery Framework (TCDF) \cite{2nauta2019causal}, which employed temporal convolutional networks (TCN) alongside a causal validation algorithm to discover causal relationships. Addressing challenges in irregularly sampled time series, Cheng et al. \cite{9cheng2023cuts} proposed  CUTS, capable of uncovering Granger causality even when data contains random missing points or non-uniform sampling intervals. To further enhance this capability, Cheng et al. \cite{5cheng2024cuts+} proposed CUTS+, which incorporated a coarse-to-fine causal discovery strategy and a message-passing graph neural network to improve the performance of causal discovery and address limitations such as large causal graphs and redundant data prediction in CUTS. Marcinkevivcs et al. \cite{4marcinkevivcs2021interpretable} proposed generalized vector autoregression (GVAR), grounded in a self-explaining neural network architecture, offering both effective causal discovery and enhanced model interpretability. Suryadi et al. \cite{10suryadi2023granger} proposed JGC, which utilized the Jacobian matrix to measure variable importance, and introduced a variable selection procedure based on significance and consistency criteria to identify Granger causality. Li et al. \cite{8li2023causal} proposed the Causal Recurrent Variational Autoencoder (CR-VAE), which integrated variational autoencoders with recurrent neural networks and incorporated sparse regularization to uncover causal relationships.

\section{Methodology}
\subsection{Time series prediction}
As Granger causality is fundamentally grounded in prediction, time series forecasting models are commonly utilized to construct causal discovery frameworks. In this study, to maintain simplicity and computational efficiency, we adopt a simple MLP to build our model, which can be formulated as follows:
\begin{equation}
\begin{split}
     \hat{x}_{t,1:D}&=f_{\theta}(x_{t-k:t-1,1:D})\\
                    &=W_2\sigma(W_1x_{t-k:t-1,1:D}+b_1)+b_2
\end{split}
\end{equation}
where $f_{\theta}$ represents the MLP parameterized by $\theta$, $W_1\in \mathbb{R}^{H\times D}$ and $W_2\in \mathbb{R}^{D\times H}$ are weights, $b^1\in \mathbb{R}^H$ and $b^2 \in\mathbb{R}^{D}$ are bias, $\sigma$ is the ReLU activation function, $H$ denotes the hidden layer size.

Then, the predicted loss is defined as:
\begin{equation}
   \mathcal{L}_{pred}=\frac{1}{T-k}\sum_{t=k+1}^{T}( \hat{x}_{t,1:D}-x_{t-k:t-1,1:D})^{2}
\end{equation}

\subsection{Gradient-based causal discovery}
After constructing the time series forecasting model, a novel gradient-based approach is proposed to extract causal relationships from neural networks instead of the traditional first-layer weights. 

The intuition behind gradient-based causal discovery is that the neural network's input-output gradient $\frac{\partial \hat{x}_{t,i}}{\partial x_{t-k:t-1,j}}$ quantifies the sensitivity of the network's prediction $\hat{x}_{t,i}$ with respect to the input past values $x_{t-k:t-1,j}$. A large gradient magnitude indicates that perturbing the past value of $x_j$ produces a substantial change in the predicted $x_i$, aligning with the Granger causality principle that \textit{causal influence from $x_j$ to $x_i$ exists when the historical information of $x_j$ significantly affects the prediction of $x_i$}. Conversely, a near-zero gradient suggests that $x_i$ is insensitive to $x_j$, indicating the absence of a causal relationship.

Specifically, for each time series $i$, the predicted values across all time steps are summed into a scalar function, which is denoted as:
\begin{equation}
   s_{i}=\sum_{t=k+1}^{T}\hat{x}_{t,i}
\end{equation}

Subsequently, for each scalar function $s_{i}$, the gradient is computed with respect to every element of the input time series $x_{t-k:t-1,1:D}$, which is denoted as:
\begin{equation}
\begin{split}
    g^{i} &=\nabla_{x_{t-k:t-1,1:D}} s_{i}\\
          &=\left[\begin{array}{ccc}
\frac{\partial s_{i}}{\partial x_{t-k,1}}         &\cdots    & \frac{\partial s_{i}}{\partial x_{t-k,D}} \\
\vdots                         & \vdots   &       \vdots                      \\
\frac{\partial s_{i}}{\partial x_{t-1,1}}    & \cdots   & \frac{\partial s_{i}}{\partial x_{t-1,D}}
\end{array}\right] \in \mathbb{R}^{k\times D}
\end{split}
\end{equation}

The $j$-th column of $g^i$ represents the gradient of the prediction $\hat{x}_{t,i}$ with respect to the past value of $x_j$ across all time steps $t$, which is denoted as:
\begin{equation} 
g^{i}_{:,j}= \left( \frac{\partial s_i}{\partial x_{t-1,j}}, \frac{\partial s_i}{\partial x_{t-2,j}}, \dots, \frac{\partial s_i}{\partial x_{t-k,j}} \right)^\top \in\mathbb{R}^{k}
\end{equation}

To obtain a robust measure of causal strength, we take the element-wise absolute value of the gradients and average over time:
\begin{equation}
GC_{i,j} =\frac{1}{k}\sum_{\tau=1}^{k}|g^{i}_{\tau,j}|
\end{equation}
where $g^{i}_{\tau,j}=\frac{\partial s_i}{\partial x_{t-\tau,j}}$ reflects the instantaneous sensitivity of the predicted $\hat{x}_{t,i}$ to the past value of $x_j$ at time step $t-\tau$, $\left|\cdot\right|$ represents the absolute value, $GC_{i,j}$ is the $i^{th}$ row and $j^{th}$ column of the discovered causal matrix $GC$. 

The absolute value ensures that both positive and negative contributions are counted as evidence of influence, as the sign of the gradient may vary over time due to oscillations or nonlinear interactions. Averaging the gradients over time mitigates the influence of noise and short-term fluctuations, providing a more robust estimate of causal strength.

To obtain the sparse causal matrix, the $\ell_1$ regularization is imposed on $GC$ to sparse its elements. The corresponding sparsity loss is defined as:
\begin{equation}
    \mathcal{L}_{sparse}=\lambda\|GC\|_{1}
\end{equation}
where $\|\cdot\|_{1}$ is denoted as the $\ell_1$ regularization, $\lambda>0$ is the hyperparameter that controls the regularization strength.

Finally, the loss function is defined as:
\begin{equation}
\label{lossfunction}
        \mathcal{L}=\mathcal{L}_{pred}+\mathcal{L}_{sparse}\\
\end{equation}

\subsection{Post-hoc statistical tests}
In a causal matrix, a non-zero causal strength from $x$ to $y$ indicates that $x$ exerts a causal influence on $y$. Conversely, if $x$ does not cause $y$, the causal strength should ideally be zero. However, in practice, time series derived from real-world systems often contain noises. As a result, pairs of variables without a true causal relationship may still exhibit small but non-zero causal strength \cite{9-zhou2025information}. Consequently, the additional statistical test is necessary to assess whether the observed causal strength is significantly greater than zero.

Here, we extend upon the time-shift surrogate method introduced by Zhou et al \cite{8-zhou2022causality}. Specifically, their approach has two limitations: (1) causal dependence is disrupted solely through time shifting, which may be insufficient in signals with strong periodic patterns, as time shifting cannot fully decorrelate the signals and often leaves periodic components partially aligned; (2) the problem of multiple comparisons is not considered, leading to potential false positives in causal matrices. To overcome these limitations, we propose a Phase-randomization Surrogate Statistical Test (PSST) module. PSST employs the phase-randomized instead of time-shift surrogate to eliminate dependencies between variables more effectively \cite{10-rath2009surrogates}. Then, the model is trained on multiple surrogate datasets to construct an empirical null distribution for each potential causal connection. Finally, $p$-values are adjusted using the Benjamini-Hochberg false discovery rate (FDR) \cite{11-chung2007detection}. The detailed procedures are as follows:

\paragraph{Surrogate generation (phase randomization)}
For each original univariate time series \(x_{t,i}\), the Fast Fourier Transform (FFT) is computed to convert the signal from the time domain to the frequency domain:
\begin{equation}
X_i(f) = \mathcal{F}(x_{t,i}), \quad f = 0,\dots,T-1.
\end{equation}
where $\mathcal{F}$ is denoted as FFT. Then, generate a phase-randomized surrogate by replacing the phase of \(X_i(f)\) with independent uniform phases while preserving the amplitude spectrum:
\begin{equation}
\label{11}
    \widetilde{X}_i(f) = |X_i(f)| \, e^{j\,\phi_i^{\text{rand}}(f)}, \qquad \phi_i^{\text{rand}}(f)\sim \mathrm{Unif}[0,2\pi)
\end{equation}

The surrogate univariate time series $\widetilde{x}_{t,i}$ is obtained by inverse FFT:
\begin{equation}
\label{12}
\widetilde{x}_{t,i} = \mathcal{F}^{-1}\{\widetilde{X}_i(f)\}
\end{equation}
where $\mathcal{F}^{-1}$ is the inverse FFT. All $D$ univariate time series constitute the complete surrogate multivariate time series:
\begin{equation}
\label{13}
    \widetilde{x}_{t} =\{\widetilde{x}_{t,i}\}_{i=1}^{D}
\end{equation}

\paragraph{Train model on surrogate time series}
Generate \(N\) surrogate datasets \(\{\widetilde{x}_{t}^{(k)}\}_{k=1}^{N}\). For each surrogate dataset, train GCD and extract the corresponding causal matrix $\widetilde{GC}$:
\begin{equation}
    \widetilde{GC}^{k}\in \mathbb{R}^{D\times D}, \qquad k=1,\dots,N
\end{equation}

\paragraph{Edge-wise empirical null distributions}
For each candidate directed edge \((m,n)\) (from variable \(m\) to \(n\)), collect the surrogate causal strength to form an empirical null distribution:
\begin{equation}
\label{15}
    \mathcal{ND}_{mn} = \big\{\widetilde{GC}^{1}_{mn},\widetilde{GC}^{2}_{mn},\dots,\widetilde{GC}^{k}_{mn},\dots, \widetilde{GC}^{N}_{mn}\big\}
\end{equation}

\paragraph{Empirical \(p\)-value}
Let \(GC_{mn}\) denote the inferred causal strength for edge \((m,n)\) from the original $x_t$. The empirical \(p\)-value is computed as:
\begin{equation}
\label{16}
p_{mn} =\frac{1}{N}\sum_{k=1}^{N}\mathbb{I}\!\big[\widetilde{GC}^k_{mn} \ge GC_{mn}\big]
\end{equation}
where \(\mathbb{I}(\cdot)\) is the indicator function. 

%Equation~\eqref{eq:empirical-p} gives the proportion of surrogate samples under \(H_0\) that are at least as extreme as the observed value.

\paragraph{Multiple-comparison correction}
Collect all \(p\)-values \(\{p_{mn}\}\) for the \(D\times D\) candidate edges and apply the Benjamini-Hochberg FDR procedure to obtain adjusted \(p\)-values \(p_{mn}^{\mathrm{adj}}\). For the significance level $\alpha=0.05$, edge \((m,n)\) is significant if:
\[
p_{mn}^{\mathrm{adj}} < 0.05
\]
The final causal matrix $A$ is defined as:
\begin{equation}
A_{mn} = 
\begin{cases}
GC_{mn}, & p_{mn}^{\mathrm{adj}} < 0.05\\[4pt]
0, & \text{otherwise}
\end{cases}
\end{equation}

\begin{algorithm}[h]
\label{algorithm1}
\DontPrintSemicolon
\SetAlgoLined
\KwIn {Time series $x_t$; Number of surrogates $N$; Hyperparameter $\lambda$; Learning rate $lr$.}
\KwOut {Causal matrix $A$.}
Train GCD with $\lambda$ and $lr$ and compute original causal matrix $GC$ on $x_t$ using Eq.\ref{lossfunction}.\;
\For{$k = 1$ \KwTo $N$}{
Generate surrogate multivariate time series $\widetilde{x}_t^{(k)}$ using Eq.\ref{11}, \ref{12}, \ref{13}.\;
Train GCD with $\lambda$ and $lr$ and compute surrogate causal matrix $\widetilde{GC}^{k}$ on $\widetilde{x}_t^{(k)}$ using Eq.\ref{lossfunction}.\;
}
\For{$m = 1$ \KwTo $D$}{
    \For{$n = 1$ \KwTo $D$}{
        Construct the edge-wise empirical null distributions $\mathcal{ND}_{mn}$ using Eq.\ref{15}.\;
        Compute the corresponding empirical $p$-value $p_{mn}$ using Eq.\ref{16}.\;
    }
}
Apply the Benjamini-Hochberg FDR procedure to all $p$-values to obtain the adjusted $p$-values $p^{adj}$.\;
\For{$m = 1$ \KwTo $D$}{
    \For{$n = 1$ \KwTo $D$}{
        \If{$p^{adj}_{mn} < 0.05$}{
            $A_{mn} \gets GC_{mn}$\;
        }
        \Else{
            $A_{mn} \gets 0$\;
        }
    }
}
\Return{$A$}
\caption{Completion of causal matrix.}
\end{algorithm}

\section{Numerical Experiment}
\label{experiment}
In this section, we compare GCD with seven state-of-the-art models, including cMLP \& cLSTM \cite{1tank2021neural}, TCDF \cite{2nauta2019causal}, eSRU \cite{3khanna2019economy}, GVAR \cite{4marcinkevivcs2021interpretable}, CR-VAE \cite{8li2023causal}, and CUTS+ \cite{5cheng2024cuts+}, across three widely used benchmark datasets: DREAM4 in Silico Network Challenge (DREAM4) \cite{11madar2010dream3}, Lorenz-96 \cite{9karimi2010extensive}, and CausalTime \cite{12chengcausaltime}. All experiments are conducted on a Windows server equipped with an Intel Xeon Silver 4310 CPU and an NVIDIA A40 GPU (48GB RAM). 

\begin{table*}[h]
  \centering
      \caption{AUROC and AUPRC of the Lorenz-96 dataset, $D=40$, $T=1000$. We highlight the best and the second best in bold and with underlining, respectively.}
    \resizebox{1\textwidth}{!}{
    \begin{tabular}{ccccccccc}
    \toprule
    \toprule
    \multirow{3}{*}{Models} & \multicolumn{2}{c}{$F=10$}  & &  \multicolumn{2}{c}{$F=20$} & & \multicolumn{2}{c}{$F=40$}\\
     \cmidrule{2-3} \cmidrule{5-6} \cmidrule{8-9}
    & AUROC & AUPRC  & & AUROC  & AUPRC  & & AUROC  & AUPRC \\
    \midrule
        cMLP            &  0.995\smallgrey{$\pm$0.003} & 0.992\smallgrey{$\pm$0.003} & & 0.993\smallgrey{$\pm$0.004} & 0.978\smallgrey{$\pm$0.004} & & 0.972\smallgrey{$\pm$0.005} & 0.943\smallgrey{$\pm$0.006}   \\
        cLSTM           & 0.983\smallgrey{$\pm$0.008} & 0.961\smallgrey{$\pm$0.012} & & 0.932\smallgrey{$\pm$0.008} & 0.904\smallgrey{$\pm$0.009} & & 0.876\smallgrey{$\pm$0.007} & 0.839\smallgrey{$\pm$0.011}   \\
        TCDF            & 0.852\smallgrey{$\pm$0.036} & 0.733\smallgrey{$\pm$0.041} & & 0.728\smallgrey{$\pm$0.027} & 0.587\smallgrey{$\pm$0.033} & & 0.585\smallgrey{$\pm$0.046} & 0.439\smallgrey{$\pm$0.052}   \\
        eSRU            & \bf 1.000\smallgrey{$\pm$0.000} & \bf 1.000\smallgrey{$\pm$0.000} & & \underline{0.996\smallgrey{$\pm$0.001}} & \underline{0.991\smallgrey{$\pm$0.002}} & & 0.988\smallgrey{$\pm$0.003} & 0.954\smallgrey{$\pm$0.005}   \\
        GVAR            & \underline{0.996\smallgrey{$\pm$0.002}} & \underline{0.994\smallgrey{$\pm$0.003}} & & 0.992\smallgrey{$\pm$0.002} & 0.983\smallgrey{$\pm$0.004} & & 0.975\smallgrey{$\pm$0.005} & 0.946\smallgrey{$\pm$0.008}   \\
        CR-VAE          & 0.876\smallgrey{$\pm$0.022} & 0.761\smallgrey{$\pm$0.035} & & 0.755\smallgrey{$\pm$0.036} & 0.640\smallgrey{$\pm$0.037} & & 0.639\smallgrey{$\pm$0.032} & 0.528\smallgrey{$\pm$0.047}   \\
        CUTS+           & \bf 1.000\smallgrey{$\pm$0.000} & \bf 1.000\smallgrey{$\pm$0.000} & & \bf 1.000\smallgrey{$\pm$0.000} & \bf 1.000\smallgrey{$\pm$0.000} & & \underline{0.991\smallgrey{$\pm$0.002}} & \underline{0.967\smallgrey{$\pm$0.003}}   \\
        \textbf{GCD}  & \bf 1.000\smallgrey{$\pm$0.000} & \bf 1.000\smallgrey{$\pm$0.000} & & \bf 1.000\smallgrey{$\pm$0.000} & \bf 1.000\smallgrey{$\pm$0.000} & & \bf0.997\smallgrey{$\pm$0.001} & \bf 0.975\smallgrey{$\pm$0.004}      \\
    \bottomrule
    \bottomrule
    \end{tabular}}%
  \label{tab-lorenz}%
\end{table*}%

\begin{table*}[h]
      \caption{AUROC and AUPRC of DREAM4 dataset, $D=100$, $T=210$. We highlight the best and the second best in bold and with underlining, respectively.}
  \centering
    \resizebox{1\textwidth}{!}{
    \begin{tabular}{cccccccccccc}
    \toprule
    \toprule
    \multirow{2}{*}{Models} &   \multicolumn{5}{c}{AUROC}  & &   \multicolumn{5}{c}{AUPRC}\\
     \cmidrule{2-6}      \cmidrule{8-12}
     & Gene-1 & Gene-2 & Gene-3 & Gene-4 & Gene-5 & & Gene-1 & Gene-2 & Gene-3 & Gene-4 & Gene-5  \\
    \midrule
        cMLP            & 0.652 & 0.522 & 0.509 & 0.511 & 0.531  & & 0.017 & 0.015 & 0.022 & 0.025 & 0.018\\
        cLSTM           & 0.633 & 0.509 & 0.498 & 0.524 & 0.552  & & 0.035 & 0.044 & 0.037 & 0.041 & 0.033\\
        TCDF            & 0.598 & 0.491 & 0.467 & 0.567 & 0.565  & & 0.011 & 0.013 & 0.014 & 0.012 & 0.014 \\
        eSRU            & 0.647 & 0.554 & 0.545 & 0.561 & 0.559  & & 0.045 & 0.041 & 0.043 & 0.051 & 0.059 \\
        GVAR            & 0.662 & 0.569 & 0.565 & 0.578 & 0.554  & & \underline{0.103} & 0.054 & 0.082 & 0.077 & 0.065 \\
        CR-VAE          & 0.583 & 0.534 & 0.536 & 0.529 & 0.545  & & 0.025 & 0.039 & 0.028 & 0.044 & 0.057 \\  
        CUTS+           & \underline{0.738} & \underline{0.622} & \underline{0.591} & \underline{0.584} & \underline{0.594}  & & 0.086 & \underline{0.059} & \underline{0.107} & \underline{0.087} & \underline{0.071}\\
        \textbf{GCD}  & \textbf{0.782}  & \textbf{0.718}  & \textbf{0.625}  & \textbf{0.633} & \textbf{0.701} &  & \textbf{0.123}  & \textbf{0.072}  & \textbf{0.138}  & \textbf{0.109} & \textbf{0.093}\\
    \bottomrule
    \bottomrule
    \end{tabular}}%
  \label{tab-dream3}%
\end{table*}%

Since all the studies corresponding to the models compared in this work used the Area Under the Receiver Operating Characteristic Curve (AUROC) and the Area Under the Precision-Recall Curve (AUPRC) for performance evaluation, we also adopt these two metrics in the numerical simulation section to ensure consistency and comparability with prior research. In the DREAM4 analysis, the evaluation is restricted to the off-diagonal elements of the causal matrix, as self-causal relationships are not included in the ground truth structure. For the CausalTime dataset, the evaluation is confined to the upper-left quadrant of the causal matrix (see the study by Cheng et al. \cite{12chengcausaltime} for more details).

The configurations and hyperparameters of GCD are detailed in the Supplementary Materials. For the comparative models, we employ the optimal hyperparameters provided by Khanna et al. \cite{3khanna2019economy} and Zhou et al. \cite{15-zhou2024jacobian}, which comprehensively cover all considered models.

\subsection{Lorenz-96}
Lorenz-96 is formulated as the following ordinary differential equation:
\begin{equation}
\frac{\partial x_{t,i}}{\partial t} =-x_{t,i-1}\left( x_{t,i-2}-x_{t,i+1}\right)-x_{t,i}+F
\end{equation}
where $F$ represents the forcing term applied to the system, $i=1,2,\ldots,D$, $D$ is the dimension of time series. For each time series $i$, the corresponding $x_{t,i}$ is influenced by four variables: \( i-2 \), \( i-1 \), \( i \), and \( i+1 \). To investigate the model's performance on high-dimensional time series, we set \( D=40 \), \( F=\{10, 20, 40\} \), and \( T=1000 \).

Table \ref{tab-lorenz} presents the performance of each model. When $F=10$, all models except TCDF and CR-VAE effectively infer causal relationships, with GCD, eSRU, and CUTS+ achieving the best scores (AUROC and AUPRC = 1.0). As $F$ increases to 20 and 40, the time series exhibit more pronounced chaotic dynamics and heightened nonlinearity, making causal inference more challenging. Under these conditions, GCD consistently attains the highest AUROC and AUPRC, demonstrating its robustness and effectiveness on the Lorenz-96 dataset.

\subsection{DREAM4}
The DREAM4 dataset comprises five sub-datasets corresponding to five different gene regulatory networks. Each sub-dataset contains 100 time series, with 21 sampling points per series, and is replicated 10 times, resulting in a total of 210 sampling points. 

The AUROC and AUPRC results of the DREAM4 dataset are summarized in Table \ref{tab-dream3}. GCD consistently achieves the highest AUROC across all five sub-datasets, demonstrating its superior performance. Notably, the causal structures of DREAM4 are extremely sparse, with the proportions of true causal variables in the five sub-datasets being only 1.77\%, 2.51\%, 1.96\%, 2.13\%, and 1.94\%, respectively. Considering AUPRC is particularly sensitive to class imbalance, it effectively reflects the model’s ability to detect such sparse causal relationships. Therefore, the high AUPRC performance of GCD highlights its strong capability in uncovering sparse causal relationships from high-dimensional time series.

\subsection{CausalTime}
CausalTime is a benchmark dataset recently introduced by Cheng et al. \cite{12chengcausaltime}, comprising three sub-datasets: AQI ($D=72$), Traffic ($D=40$), and Medical ($D=40$). The time series length of all sub-dataset is 19200. 

\begin{table*}[h]
\caption{AUROC and AUPRC of the CausalTime dataset, $D=72/40/40$, $T=19200$. We highlight the best and the second best in bold and with underlining, respectively.}
  \centering
    \resizebox{1\textwidth}{!}{
    \begin{tabular}{ccccccccc}
    \toprule
    \toprule
    \multirow{3}{*}{Models} & \multicolumn{2}{c}{ AQI }  & &  \multicolumn{2}{c}{Traffic} & & \multicolumn{2}{c}{Medical}\\
     \cmidrule{2-3} \cmidrule{5-6} \cmidrule{8-9}
    & AUROC  & AUPRC& & AUROC & AUPRC & & AUROC  & AUPRC \\
    \midrule
        cMLP          & 0.645\smallgrey{$\pm$0.013} & 0.656\smallgrey{$\pm$0.013} & & 0.551\smallgrey{$\pm$0.028} & 0.256\smallgrey{$\pm$0.031} & & 0.574\smallgrey{$\pm$0.009} & 0.463\smallgrey{$\pm$0.012}   \\
        cLSTM         & 0.717\smallgrey{$\pm$0.007} & 0.717\smallgrey{$\pm$0.006} & & 0.603\smallgrey{$\pm$0.005} & 0.358\smallgrey{$\pm$0.049} & & 0.531\smallgrey{$\pm$0.015} & 0.426\smallgrey{$\pm$0.025}   \\
        TCDF          & 0.414\smallgrey{$\pm$0.020} & 0.652\smallgrey{$\pm$0.008} & & 0.503\smallgrey{$\pm$0.004} & 0.363\smallgrey{$\pm$0.008} & & 0.633\smallgrey{$\pm$0.038} & 0.554\smallgrey{$\pm$0.031}   \\
        eSRU          & 0.822\smallgrey{$\pm$0.031} & 0.722\smallgrey{$\pm$0.031} & & 0.598\smallgrey{$\pm$0.019} & 0.488\smallgrey{$\pm$0.031} & & 0.756\smallgrey{$\pm$0.036} & \underline{0.735\smallgrey{$\pm$0.060}}   \\
        GVAR          & 0.643\smallgrey{$\pm$0.026} & 0.657\smallgrey{$\pm$0.047} & & 0.536\smallgrey{$\pm$0.028} & 0.379\smallgrey{$\pm$0.047} & & 0.688\smallgrey{$\pm$0.026} & 0.650\smallgrey{$\pm$0.045}   \\
        CR-VAE        & 0.675\smallgrey{$\pm$0.011} & 0.720\smallgrey{$\pm$0.026} & & 0.605\smallgrey{$\pm$0.036} & 0.376\smallgrey{$\pm$0.035} & & 0.556\smallgrey{$\pm$0.021} & 0.589\smallgrey{$\pm$0.015}   \\
        CUTS+         & \bf 0.892\smallgrey{$\pm$0.021} & \bf 0.798\smallgrey{$\pm$0.087} & & \underline{0.617\smallgrey{$\pm$0.075}} &  \underline{0.636\smallgrey{$\pm$0.119}} & & \underline{0.820\smallgrey{$\pm$0.017}} & 0.548\smallgrey{$\pm$0.134}   \\ 
        \textbf{GCD}  & \underline{0.872\smallgrey{$\pm$0.013}} & \underline{0.732\smallgrey{$\pm$0.034}} & & \bf 0.773\smallgrey{$\pm$0.015} &\bf0.639\smallgrey{$\pm$0.014} & & \bf 0.919\smallgrey{$\pm$0.014} & \bf 0.904\smallgrey{$\pm$0.019}      \\
    \bottomrule
    \bottomrule
    \end{tabular}}%
  \label{tab-causaltime}%
\end{table*}%

As presented in Table \ref{tab-causaltime}, GCD attains the highest performance on the Traffic and Medical datasets, outperforming all baseline models. For the AQI dataset, its performance is marginally lower than that of CUTS+. Notably, GCD is the only model achieving both AUROC and AUPRC values exceeding 0.9 on the Medical dataset. These results collectively demonstrate the consistent reliability and robustness of GCD across all datasets, with all evaluation metrics substantially exceeding the baseline threshold of 0.5.

\section{Applications to Complex Industrial Systems}
We evaluate the model’s performance on three complex industrial processes, including Tennessee-Eastman (TE), Ultra-processed Food (UF), and the Debutanizer (DE). Four metrics are used for evaluation: Structural Hamming Distance (SHD), F1-score, Recall, and Precision. Detailed descriptions and discussions of the experimental results are presented in the following subsections.

\subsection{Tennessee-Eastman process}
\begin{figure}[h]
\centering
\includegraphics[width=1\columnwidth]{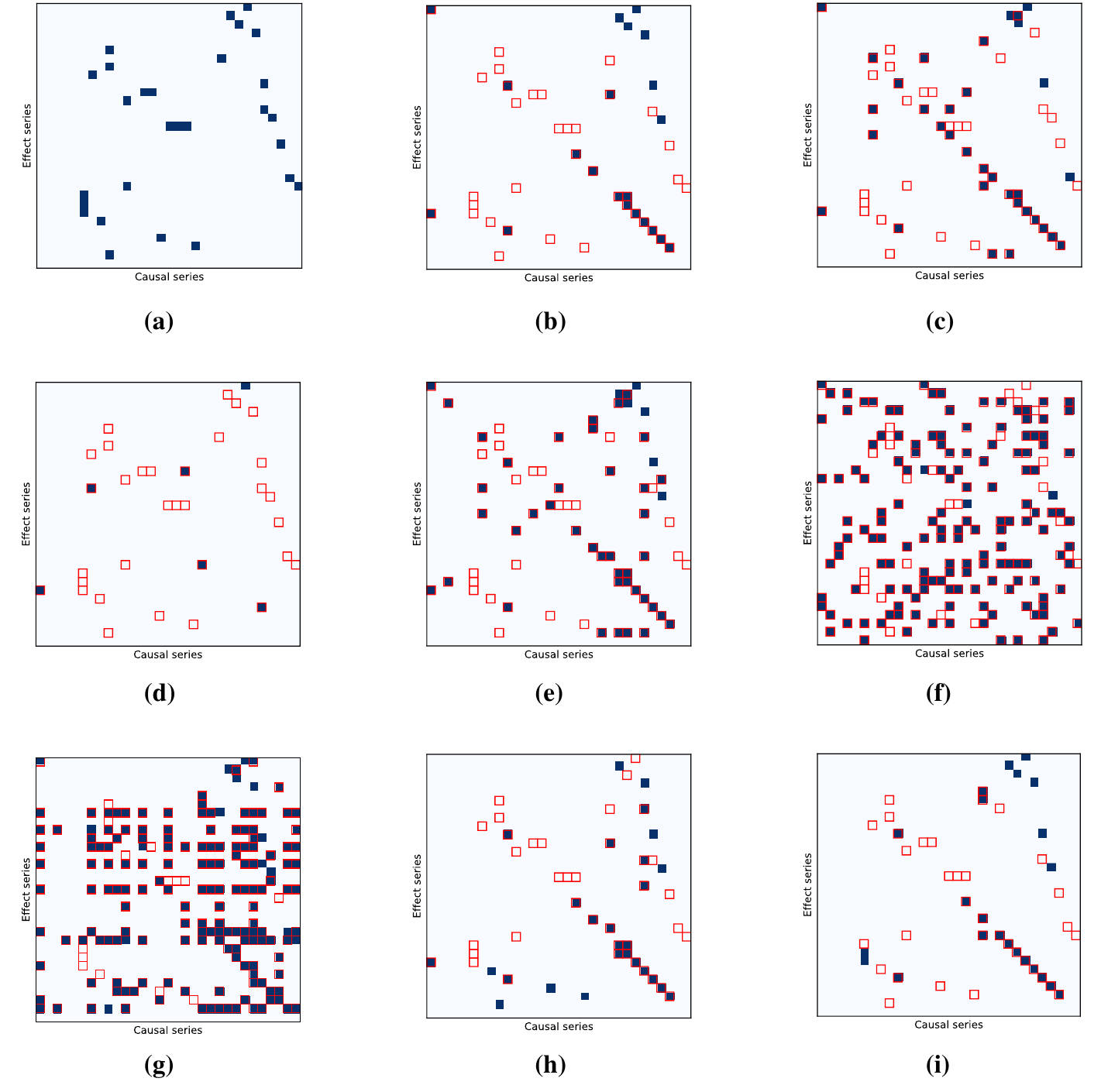} % Reduce the figure size so that it is slightly narrower than the column.
\caption{The causal discovery results of each model on TE dataset. (a) Ground truth. (b) cMLP. (c) cLSTM. (d) TCDF. (e) eSRU. (f) GVAR. (g) CR-VAE. (h) CUTS+. (i) GCD.}
\label{fig-TE}
\end{figure}

The TE process is a representative industrial chemical production system that comprises reaction, separation, and recycle units \cite{12-mcavoy1994base}. In this study, we employ the TE dataset provided by Menegozzo et al. \cite{13-menegozzo2022cipcad}, which comprises 31 variables and 1,499 time samples. The causal discovery results obtained from each model are summarized in Table \ref{tab-TE} and Fig.\ref{fig-TE}.

\begin{table}[h]
      \caption{Comparison results of the TE process, $D=31$, $T=1499$. We highlight the best and the second best in bold and with underlining, respectively.}
  \centering
    \resizebox{1\columnwidth}{!}{
    \begin{tabular}{ccccc}
    \toprule
    \toprule
    Models & SHD & F1 & Recall & Precision\\
    \midrule
        cMLP            & \underline{37}  & 0.245   &  0.214  & 0.286\\   %lam=0.05,ridge=0.01,lr=0.01
        cLSTM           & 52  & 0.161   &  0.179  & 0.147 \\    
        TCDF            & 42  & 0.059   &  0.036  & 0.167 \\
        eSRU            & 63  & 0.160   &  0.209  & 0.128 \\
        GVAR            & 174 & 0.033   &  0.107  & 0.020 \\
        CR-VAE           & 171 & 0.141   &  \bf 0.500  & 0.082 \\
        CUTS+           & 39  & \underline{0.291}   &  \underline{0.286}  & \underline{0.296}\\
        \textbf{GCD}    & \bf 35  & \bf 0.314   &  \underline{0.286}  & \bf 0.348  \\
    \bottomrule
    \bottomrule
    \end{tabular}}%
  \label{tab-TE}%
\end{table}%

As shown in Fig.\ref{fig-TE}, the causal matrices discovered by GCD, cMLP, cLSTM, eSRU, and CUTS+ are most consistent with the ground truth, exhibiting a clear, sparse pattern with only a few errors. Moreover, GCD achieves the best SHD, precision, and recall, indicating its effectiveness in capturing causal relationships while suppressing spurious connections. In constract, GVAR and CR-VAE discover numerous spurious connections, resulting in an overly dense causal matrix. TCDF captures only part of the true causal structure while omitting many true causal relationships. Overall, GCD discovered the most accurate causal matrix among all approaches, more aligned with the true causal relationships in the TE process.

\subsection{Ultra-processed food process}
\begin{figure}[h]
\centering
\includegraphics[width=1\columnwidth]{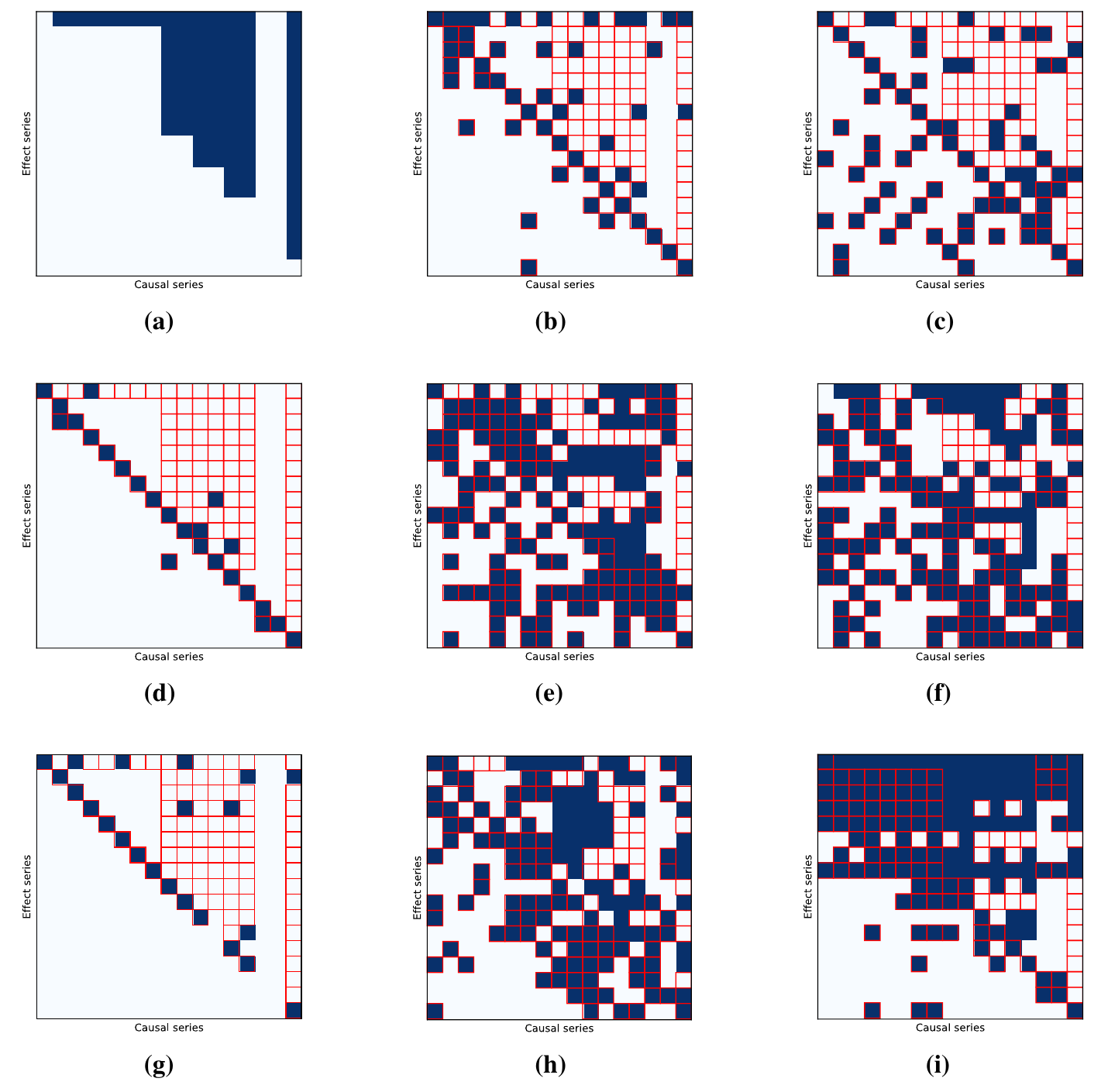} % Reduce the figure size so that it is slightly narrower than the column.
\caption{The causal discovery results of each model on UF dataset. (a) Ground truth. (b) cMLP. (c) cLSTM. (d) TCDF. (e) eSRU. (f) GVAR. (g) CR-VAE. (h) CUTS+. (i) GCD.}
\label{fig-UP}
\end{figure}

The UF factory aims to minimize the inherent variability of natural raw materials to ensure greater product uniformity. During this industrial process, 17 sensors installed across different equipment continuously monitor operational conditions, yielding a total of 23132 data points. The causal relationships discovered by each model and the performance comparison are illustrated in Fig.\ref{fig-UP} and Table \ref{tab-UP}.

\begin{table}[h]
      \caption{Comparison results of the UF process, $D=17$, $T=23132$. We highlight the best and the second best in bold and with underlining, respectively.}
  \centering
    \resizebox{1\columnwidth}{!}{
    \begin{tabular}{ccccc}
    \toprule
    \toprule
    Models & SHD & F1 & Recall & Precision\\
    \midrule
        cMLP            & 100    & 0.254    & 0.205     & 0.333\\   %lam=0.05,ridge=0.01,lr=0.01
        cLSTM           & 122    & 0.218    & 0.205     & 0.233\\    
        TCDF            & \underline{99}     & 0.075    & 0.049     & 0.167\\
        eSRU            & 158    & 0.325    & 0.458     & 0.252\\
        GVAR            & 153    & 0.349    & 0.467     & 0.270\\
        CR-VAE           & \bf 89 & 0.152    & 0.096     & \underline{0.364}\\
        CUTS+           & 125    & \underline{0.444}    & \underline{0.602}     & 0.352\\
        \textbf{GCD}    & 113 &   \bf 0.511  & \bf 0.711 & \bf 0.399\\
    \bottomrule
    \bottomrule
    \end{tabular}}%
  \label{tab-UP}%
\end{table}%

Compared with the TE process, all models exhibit a pronounced performance degradation in the UF process. Specifically, cMLP, cLSTM, TCDF, and CR-VAE recover only parts of the ground-truth structure, while eSRU, GVAR, and CUTS+ generate many false-positive connections. In contrast, GCD more effectively captures the hierarchical and modular dependencies inherent to the UF process, attaining the lowest SHD and the highest F1-score and Precision, while ranking second in Recall among all evaluated methods.

\subsection{Debutanizer process}

\begin{figure}[h]
\centering
\includegraphics[width=1\columnwidth]{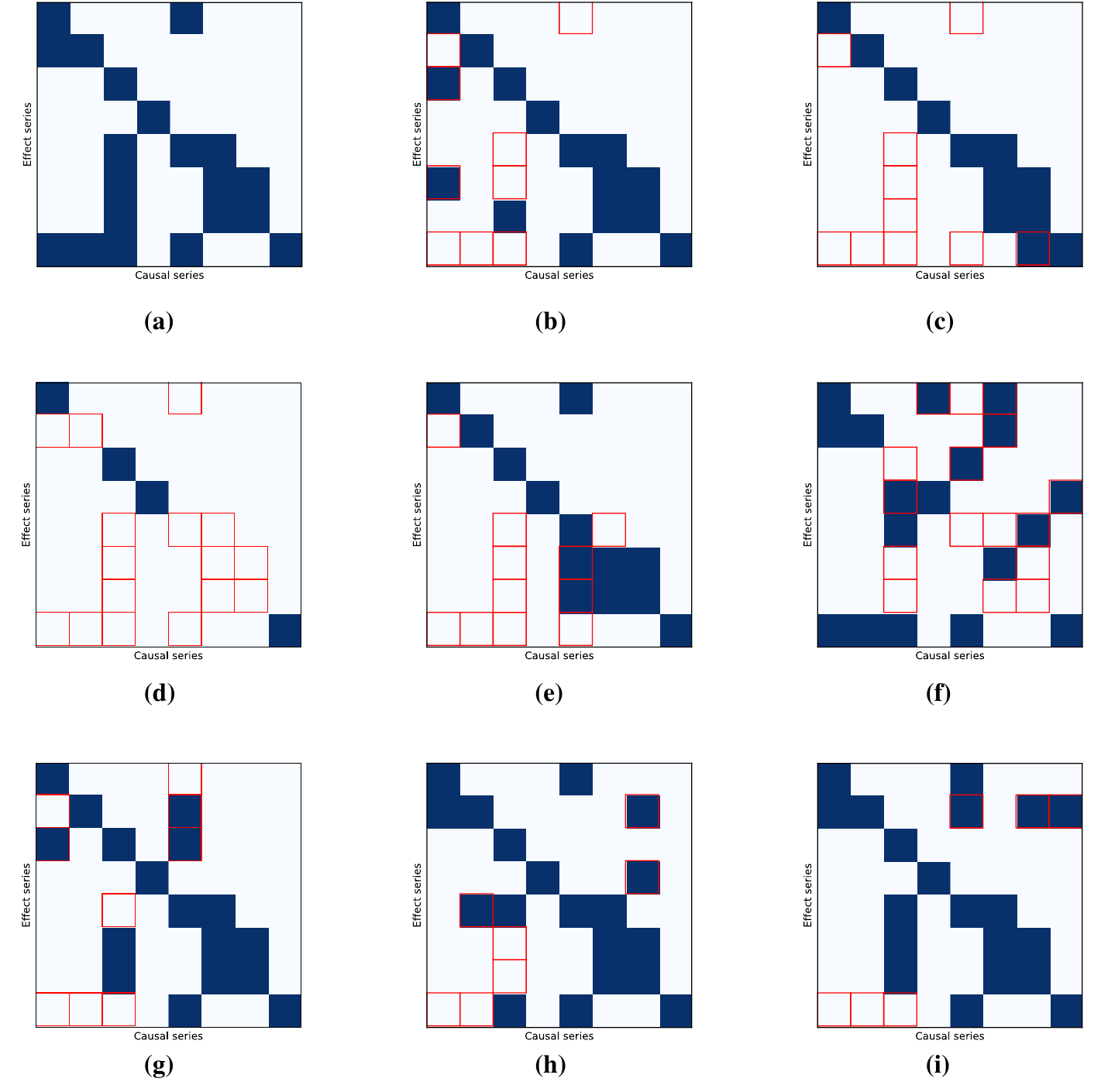} % Reduce the figure size so that it is slightly narrower than the column.
\caption{The causal discovery results of each model on the Debutanizer dataset. (a) Ground truth. (b) cMLP. (c) cLSTM. (d) TCDF. (e) eSRU. (f) GVAR. (g) CR-VAE. (h) CUTS+. (i) GCD.}
\label{fig-DE}
\end{figure}

The DE process is a distillation system widely employed in the petroleum and petrochemical industries to separate butane components from lighter and heavier hydrocarbons \cite{14-fortuna2005soft}. The dataset comprises 8 variables, each containing 2,394 sampling points, and is publicly available at \url{https://home.isr.uc.pt/~fasouza/datasets.html}. The comparative results are shown in Table \ref{tab-DE} and Fig.\ref{fig-DE}.

\begin{table}[h]
      \caption{Comparison results of the Debutanizer process process, $D=8$, $T=2394$. We highlight the best and the second best in bold and with underlining, respectively.}
  \centering
    \resizebox{1\columnwidth}{!}{
    \begin{tabular}{ccccc}
    \toprule
    \toprule
    Models & SHD & F1 & Recall & Precision\\
    \midrule
        cMLP            & 9  & 0.743 & 0.650  & 0.867 \\   %lam=0.05,ridge=0.01,lr=0.01
        cLSTM           & 10 & 0.687 & 0.550  & \underline{0.917} \\   
        TCDF            & 16 & 0.333 & 0.200  & \bf 1.000 \\ 
        eSRU            & 11 & 0.667 & 0.550  & 0.846 \\ 
        GVAR            & 16 & 0.579 & 0.550  & 0.611 \\ 
        CR-VAE           & 9  & 0.757 & 0.700  & 0.824 \\ 
        CUTS+           & \underline{8}  & \underline{0.821} & \underline{0.800}  & 0.842 \\ 
        \textbf{GCD}    & \bf 6  & \bf 0.850 & \bf 0.850  & 0.850 \\ 
    \bottomrule
    \bottomrule
    \end{tabular}}%
  \label{tab-DE}%
\end{table}%

As illustrated in Fig.\ref{fig-DE}, most models exhibit good performance on the DE process, with GCD, cMLP, CR-VAE, and CUTS+ showing only slight deviations from the ground truth. Our model achieves the best results in SHD, F1-score, and Recall, while performing marginally worse than cLTM and TCDF in Precision.

Overall, the experiments conducted across three complex industrial processes validate the effectiveness of GCD and highlight its potential to advance causal analysis in industrial process systems.

\subsection{Model complexity}

To assess model efficiency, we first analyze each model's time complexity. In this context, $e$ denotes the training epochs, $D$ represents the dimensionality of the time series, and $T$ is the time series length. $H_l$ indicates the number of neurons in layer $l$, $L$ is the total number of hidden layers, $h_r$ specifies the hidden size of recurrent neural networks (e.g., LSTM (cLSTM), SRU (eSRU), and GRU (CR-VAE, CUTS+)), $k_T$ denotes the kernel size in the TCN used in TCDF, $K_G$ represents the order of the GVAR, $s$ is the time step in CUTS+, and $N$ corresponds to the number of surrogate tests employed in GCD. The analysis results are provided in Table \ref{time-complexity}.

\begin{table}[h]
\caption{Time Complexity Analysis of Different Models.}
\label{time-complexity}
\begin{center}  
\resizebox{0.95\columnwidth}{!}{
\begin{tabular}{cc}
\toprule
\toprule
Models & Time complexity \\
\midrule
cMLP         &$\mathcal{O}(eD\sum_{l=1}^{L-1}H_lH_{l+1})$\\
cLSTM        &$\mathcal{O}(4eDT(h_{r}^2+h_{r}D))$\\
TCDF         &$\mathcal{O}(eDTk_T\sum_{l=1}^{L-1}H_lH_{l+1})$\\
eSRU         &$\mathcal{O}(eDT(h_{r}d_{\phi}+h_{r}D))$\\
GVAR         &$\mathcal{O}(2eDK_G\sum_{l=1}^{L-1}H_lH_{l+1})$\\
CR-VAE        &$\mathcal{O}(eDT\sum_{l=1}^{L-1}H_lH_{l+1}+h_{r}^2+h_{r}D)$\\
CUTS+        &$\mathcal{O}(eDTs(h_{r}^2+h_{r}D))$\\
GCD          &$\mathcal{O}(e(N+1)\sum_{l=1}^{L-1}H_lH_{l+1})+D^2$\\     
\bottomrule
\bottomrule
\end{tabular}}
\end{center}
\end{table}

The results indicate that both cMLP and GCD exhibit lower time complexity compared to other models. Specifically, cMLP has a complexity of $\mathcal{O}\bigl(eD\sum_{l=1}^{L-1} H_l H_{l+1}\bigr)$, which positive correlated with the input dimension $D$. While the primary time complexity of GCD is $\mathcal{O}\bigl(e(N+1)\sum_{l=1}^{L-1} H_l H_{l+1}\bigr)$, which depends on the number of surrogate tests $N$. In many industrial applications, the input dimension $D$ can be very large, often exceeding $N$ ($D > N+1$). Under these conditions, GCD can offer lower computational overhead and increased causal discovery performances.

Moreover, we also compare the total number of parameters, training time per epoch, and total training time across three industrial datasets. The results are provided in Table \ref{complexity}.

\begin{table}[h]
\caption{Model efficiency comparisons on the TE, UF, and DE datasets.}
\label{complexity}
\begin{center}  
\resizebox{1\columnwidth}{!}{
\begin{tabular}{cccccc}
\toprule
\toprule
Datasets & Models & \makecell{Total \\Parameter} & \makecell{Training time per \\epoch (s)} & \makecell{ Total training \\ Time (s)} \\
\midrule
\multirow{6}{*}{\makecell{TE\\ process \\($D=31$)}}                & cMLP          &486715   &0.057      &114.44     \\
                                                                   & cLSTM         &1652331  &0.142      &285.22     \\
                                                                   & TCDF          &70184    &0.105      &210.39     \\
                                                                   & eSRU          &218302   &0.133      &267.56     \\
                                                                   & GVAR          &1002610  &0.138      &130.85    \\
                                                                   & CR-VAE         &606303   &0.161      &323.80     \\
                                                                   & CUTS+         &525462   &0.082      &164.34      \\          
                                                                   & GCD           &64543    &0.036      &72.22       \\     
\midrule
\multirow{6}{*}{\makecell{UF \\process\\($D=17$)}}                 & cMLP          &73954    &0.042      &85.02      \\
                                                                   & cLSTM         &810916   &0.809      &1618.82    \\
                                                                   & TCDF          &46052    &0.055      &110.28      \\
                                                                   & eSRU          &55096    &0.180      &360.59      \\
                                                                   & GVAR          &309890   &0.653      &1307.64     \\
                                                                   & CR-VAE         &296273   &0.063      &127.21      \\
                                                                   & CUTS+         &169742   &0.055      &110.83      \\          
                                                                   & GCD           &35857    &0.034      &68.26      \\     
                                 \hline
\multirow{6}{*}{\makecell{DE \\process\\($D=8$)}}                  & cMLP          &67208    &0.022      &43.95    \\
                                                                   & cLSTM         &352808   &0.041      &82.64    \\
                                                                   & TCDF          &18112    &0.095      &191.67   \\
                                                                   & eSRU          &24496    &0.075      &151.28    \\
                                                                   & GVAR          &73640    &0.086      &172.94    \\
                                                                   & CR-VAE         &136712   &0.033      &66.68      \\
                                                                   & CUTS+         &864278   &0.035      &71.35      \\          
                                                                   & GCD           &17416    &0.017      &35.58      \\   
\bottomrule
\bottomrule
\end{tabular}}
\end{center}
\end{table}

Since GCD employs a single model rather than $D$ models (component-wise architecture) and utilizes a simple MLP for time series forecasting, it contains fewer tunable parameters than existing approaches, thereby considerably reducing computational overhead.

\subsection{Sensitivity against hyperparameter tuning}
\begin{figure*}[h]
\centering
\includegraphics[width=1\textwidth]{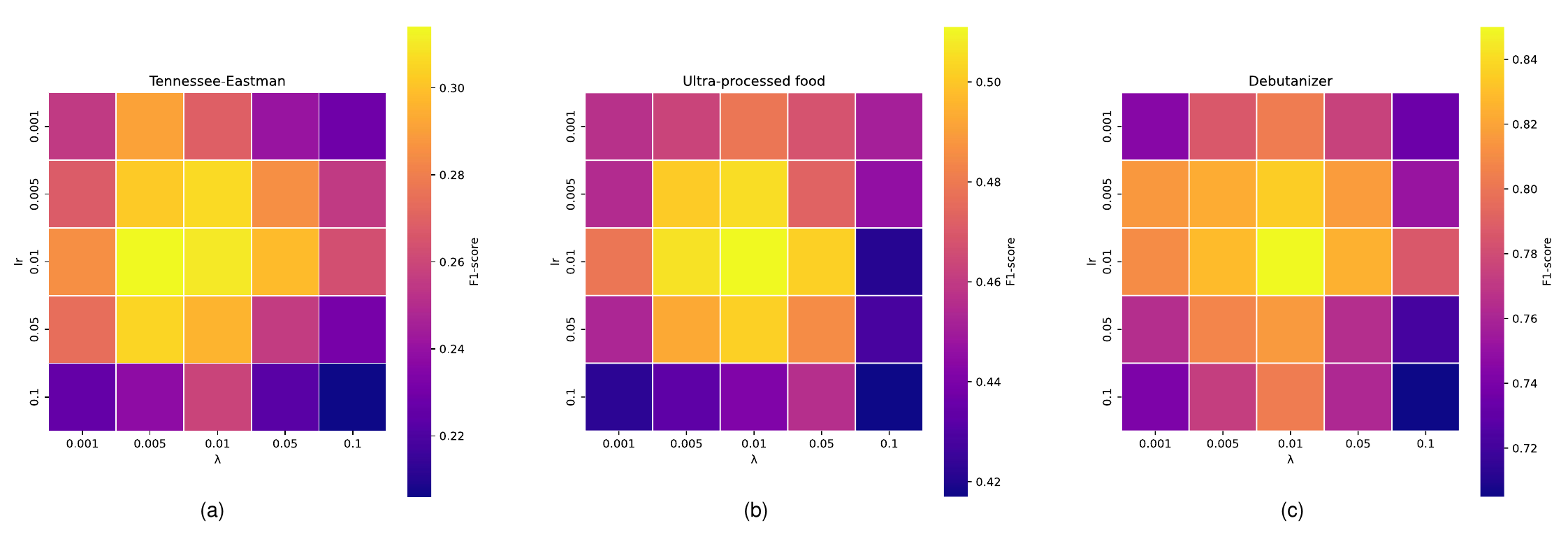} % Reduce the figure size so that it is slightly narrower than the column.
\caption{Hyperparameter tuning results. (a) TE. (b) UF. (c) DE.}
\label{fig-heatmap}
\end{figure*}

Considering GCD employs $\ell_1$ regularization to promote sparsity in the causal matrix, it is crucial to evaluate the model’s robustness and sensitivity to hyperparameter variations. Therefore, we conduct sensitivity analysis on three industrial processes. The $\ell_1$ regularization hyperparameter is varied within the range $\lambda \in [10^{-3}, 10^{-1}]$, while the learning rate is tuned over $lr \in [10^{-3}, 10^{-1}]$. F1-score is adopted as the evaluation metric. The corresponding results are presented in Fig.\ref{fig-heatmap}.

The experimental results show that GCD maintains stable performance under various hyperparameter configurations. While moderate adjustments to $\lambda$ and $lr$ cause only minimal variation in the F1-score, extreme parameter values lead to a performance decrease. These outcomes highlight the robustness of GCD to hyperparameter tuning.

\subsection{Ablation studies}

\begin{figure*}[h]
\centering
\includegraphics[width=1\textwidth]{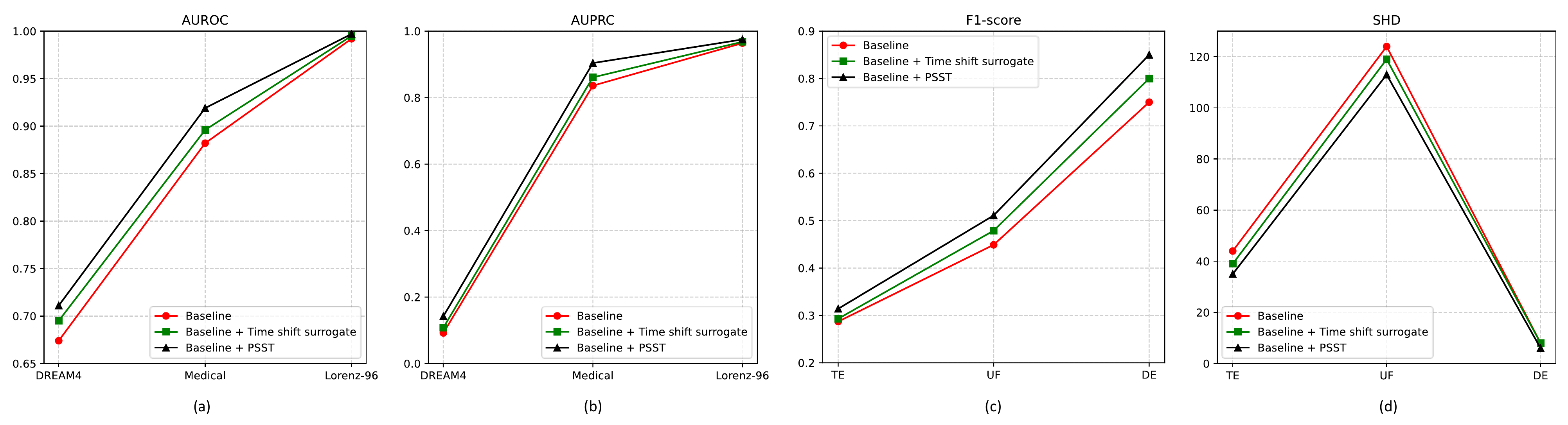} % Reduce the figure size so that it is slightly narrower than the column.
\caption{Ablation study results on numerical simulation (Lorenz-96, DREAM4, Medical) and industrial process (TE, UF, DE) datasets. (a) AUROC (numerical simulations). (b) AUPRC (numerical simulations). (c) F1-score (industrial process). (d) SHD (industrial process).}
\label{ab}
\end{figure*}

The ablation studies are designed to assess the individual contributions of each component within our model, thereby elucidating the influence of its core elements on causal discovery performance. We investigate the effect of the PSST module on overall performance. Specifically, we design three variants: baseline, baseline + traditional time-shift surrogate, and baseline + PSST module. The baseline model uses GCD. Experiments are conducted on both simulation datasets (Lorenz-96, DREAM4, Medical) and industrial processes (TE, UF, DE). The corresponding results are presented in Fig.\ref{ab}.

The results indicate that the PSST module effectively improves causal inference performance, achieving higher AUROC, AUPRC, F1-score, and SHD than the traditional time-shift surrogate method. Specifically, in three numerical simulation datasets, the PSST module improves AUROC by 3.7\%, 3.9\%, and 0.5\%, whereas the time-shift surrogate method yields respective improvements of 1.6\%, 2.3\%, and 0.3\%. Similarly, in terms of AUPRC, the PSST module achieves gains of 5.0\%, 6.8\%, and 1.1\%, outperforming the time-shift surrogate’s increases of 1.6\%, 2.5\%, and 0.4\%. In three industrial processes, the PSST module yields F1-score improvements of 2.7\%, 6.2\%, and 10\%, outperforming the time-shift surrogate, which attains increases of 0.6\%, 3.0\%, and 5.0\%. Likewise, PSST achieves larger SHD reductions (9, 11, and 2) than the time-shift surrogate method (5, 5, and 0).

\section{Conclusion}
In this study, we propose Gradient-based Causal Discovery (GCD), a novel framework for discovering causal relationships from multivariate time series. Different from previous models that rely on component-wise architectures and derive causality from the first-layer weights of neural networks, GCD employs a single MLP and extracts causal relationships from the gradients between the network’s inputs and outputs. Extensive experiments on Lorenz-96, DREAM4, and CausalTime demonstrate the superior performance of GCD over existing baselines. Evaluations on three industrial processes, including Tennessee-Eastman, Ultra-processed Food, and Debutanizer, validate the effectiveness of GCD in uncovering the underlying causal structures of complex systems while substantially reducing training time and resource consumption. These results highlight the potential of GCD as a powerful and scalable tool for causal discovery in industrial practice.

\bibliography{ref}{}

% Generated by IEEEtran.bst, version: 1.14 (2015/08/26)
\begin{thebibliography}{10}
\providecommand{\url}[1]{#1}
\csname url@samestyle\endcsname
\providecommand{\newblock}{\relax}
\providecommand{\bibinfo}[2]{#2}
\providecommand{\BIBentrySTDinterwordspacing}{\spaceskip=0pt\relax}
\providecommand{\BIBentryALTinterwordstretchfactor}{4}
\providecommand{\BIBentryALTinterwordspacing}{\spaceskip=\fontdimen2\font plus
\BIBentryALTinterwordstretchfactor\fontdimen3\font minus \fontdimen4\font\relax}
\providecommand{\BIBforeignlanguage}[2]{{%
\expandafter\ifx\csname l@#1\endcsname\relax
\typeout{** WARNING: IEEEtran.bst: No hyphenation pattern has been}%
\typeout{** loaded for the language `#1'. Using the pattern for}%
\typeout{** the default language instead.}%
\else
\language=\csname l@#1\endcsname
\fi
#2}}
\providecommand{\BIBdecl}{\relax}
\BIBdecl

\bibitem{4-zhang2024regioselective}
X.~Zhang, T.~Yan, H.~Hou, J.~Yin, H.~Wan, X.~Sun, Q.~Zhang, F.~Sun, Y.~Wei, M.~Dong \emph{et~al.}, ``Regioselective hydroformylation of propene catalysed by rhodium-zeolite,'' \emph{Nature}, vol. 629, no. 8012, pp. 597--602, 2024.

\bibitem{3-wang2023causal}
H.~Wang, R.~Liu, S.~X. Ding, Q.~Hu, Z.~Li, and H.~Zhou, ``Causal-trivial attention graph neural network for fault diagnosis of complex industrial processes,'' \emph{IEEE Transactions on Industrial Informatics}, vol.~20, no.~2, pp. 1987--1996, 2023.

\bibitem{27granger1969investigating}
C.~W. Granger, ``Investigating causal relations by econometric models and cross-spectral methods,'' \emph{Econometrica: journal of the Econometric Society}, pp. 424--438, 1969.

\bibitem{1-hua2025integrated}
D.~Hua, J.~Dong, K.~Peng, and S.~Simani, ``An integrated distributed fault diagnosis framework for large-scale industrial processes based on spatio--temporal causal analysis,'' \emph{IEEE Transactions on Industrial Informatics}, 2025.

\bibitem{6-sui2025attribution}
Q.~Sui, Y.~Wang, C.~Liu, K.~Wang, and B.~Sun, ``Attribution-aided nonlinear granger causality discovery method and its industrial application,'' \emph{IEEE Transactions on Industrial Informatics}, 2025.

\bibitem{5-runge2023causal}
J.~Runge, A.~Gerhardus, G.~Varando, V.~Eyring, and G.~Camps-Valls, ``Causal inference for time series,'' \emph{Nature Reviews Earth \& Environment}, vol.~4, no.~7, pp. 487--505, 2023.

\bibitem{liu2025spatiotemporal}
M.~Liu, C.~Yu, X.~Yang, Y.~Xu, H.~Dong, Z.~Li, Z.~Si, X.~Yang, J.~Huang, Z.~Shi \emph{et~al.}, ``A spatiotemporal causal model for revealing developmental changes in infants' brain effective connectivity networks during the first year of life,'' \emph{IEEE Transactions on Biomedical Engineering}, 2025.

\bibitem{1tank2021neural}
A.~Tank, I.~Covert, N.~Foti, A.~Shojaie, and E.~B. Fox, ``Neural granger causality,'' \emph{IEEE Transactions on Pattern Analysis and Machine Intelligence}, vol.~44, no.~8, pp. 4267--4279, 2022.

\bibitem{3khanna2019economy}
S.~Khanna and V.~Y. Tan, ``Economy statistical recurrent units for inferring nonlinear granger causality,'' in \emph{International Conference on Learning Representations}, 2019.

\bibitem{9cheng2023cuts}
Y.~Cheng, R.~Yang, T.~Xiao, Z.~Li, J.~Suo, K.~He, and Q.~Dai, ``Cuts: Neural causal discovery from irregular time-series data,'' in \emph{The Eleventh International Conference on Learning Representations}, 2023.

\bibitem{10suryadi2023granger}
S.~Suryadi, L.~Y. Chew, and Y.-S. Ong, ``Granger causality using jacobian in neural networks,'' \emph{Chaos: An Interdisciplinary Journal of Nonlinear Science}, vol.~33, no.~2, 2023.

\bibitem{5cheng2024cuts+}
Y.~Cheng, L.~Li, T.~Xiao, Z.~Li, J.~Suo, K.~He, and Q.~Dai, ``Cuts+: High-dimensional causal discovery from irregular time-series,'' in \emph{Proceedings of the AAAI Conference on Artificial Intelligence}, vol.~38, 2024, pp. 11\,525--11\,533.

\bibitem{8bussmann2021neural}
B.~Bussmann, J.~Nys, and S.~Latr{\'e}, ``Neural additive vector autoregression models for causal discovery in time series,'' in \emph{Discovery Science: 24th International Conference, DS 2021, Halifax, NS, Canada, October 11--13, 2021, Proceedings 24}.\hskip 1em plus 0.5em minus 0.4em\relax Springer, 2021, pp. 446--460.

\bibitem{2nauta2019causal}
M.~Nauta, D.~Bucur, and C.~Seifert, ``Causal discovery with attention-based convolutional neural networks,'' \emph{Machine Learning and Knowledge Extraction}, vol.~1, no.~1, p.~19, 2019.

\bibitem{4marcinkevivcs2021interpretable}
R.~Marcinkevi{\v{c}}s and J.~E. Vogt, ``Interpretable models for granger causality using self-explaining neural networks,'' in \emph{International Conference on Learning Representations (ICLR 2021)}.\hskip 1em plus 0.5em minus 0.4em\relax OpenReview, 2021.

\bibitem{8li2023causal}
H.~Li, S.~Yu, and J.~Principe, ``Causal recurrent variational autoencoder for medical time series generation,'' in \emph{Proceedings of the AAAI conference on artificial intelligence}, vol.~37, 2023, pp. 8562--8570.

\bibitem{9-zhou2025information}
W.~Zhou, S.~Bai, Y.~Xie, Y.~He, Q.~Zhao, and B.~Chen, ``An information-theoretic approach for heterogeneous differentiable causal discovery,'' \emph{Neural Networks}, vol. 188, p. 107417, 2025.

\bibitem{8-zhou2022causality}
W.~Zhou, S.~Yu, and B.~Chen, ``Causality detection with matrix-based transfer entropy,'' \emph{Information Sciences}, vol. 613, pp. 357--375, 2022.

\bibitem{10-rath2009surrogates}
C.~R{\"a}th and R.~Monetti, ``Surrogates with random fourier phases,'' in \emph{Topics on chaotic systems: selected papers from chaos 2008 international conference}.\hskip 1em plus 0.5em minus 0.4em\relax World Scientific, 2009, pp. 274--285.

\bibitem{11-chung2007detection}
P.-J. Chung, J.~F. Bohme, C.~F. Mecklenbrauker, and A.~O. Hero, ``Detection of the number of signals using the benjamini-hochberg procedure,'' \emph{IEEE Transactions on Signal Processing}, vol.~55, no.~6, pp. 2497--2508, 2007.

\bibitem{11madar2010dream3}
A.~Madar, A.~Greenfield, E.~Vanden-Eijnden, and R.~Bonneau, ``Dream3: network inference using dynamic context likelihood of relatedness and the inferelator,'' \emph{PloS one}, vol.~5, no.~3, p. e9803, 2010.

\bibitem{9karimi2010extensive}
A.~Karimi and M.~R. Paul, ``Extensive chaos in the lorenz-96 model,'' \emph{Chaos: An interdisciplinary journal of nonlinear science}, vol.~20, no.~4, 2010.

\bibitem{12chengcausaltime}
Y.~Cheng, Z.~Wang, T.~Xiao, Q.~Zhong, J.~Suo, and K.~He, ``Causaltime: Realistically generated time-series for benchmarking of causal discovery,'' in \emph{The Twelfth International Conference on Learning Representations}, 2024.

\bibitem{15-zhou2024jacobian}
W.~Zhou, S.~Bai, S.~Yu, Q.~Zhao, and B.~Chen, ``Jacobian regularizer-based neural granger causality,'' in \emph{International Conference on Machine Learning}.\hskip 1em plus 0.5em minus 0.4em\relax PMLR, 2024, pp. 61\,763--61\,782.

\bibitem{12-mcavoy1994base}
T.~McAvoy and N.~Ye, ``Base control for the tennessee eastman problem,'' \emph{Computers \& Chemical Engineering}, vol.~18, no.~5, pp. 383--413, 1994.

\bibitem{13-menegozzo2022cipcad}
G.~Menegozzo, D.~Dall’Alba, and P.~Fiorini, ``Cipcad-bench: Continuous industrial process datasets for benchmarking causal discovery methods,'' in \emph{2022 IEEE 18th International Conference on Automation Science and Engineering (CASE)}.\hskip 1em plus 0.5em minus 0.4em\relax IEEE, 2022, pp. 2124--2131.

\bibitem{14-fortuna2005soft}
L.~Fortuna, S.~Graziani, and M.~G. Xibilia, ``Soft sensors for product quality monitoring in debutanizer distillation columns,'' \emph{Control Engineering Practice}, vol.~13, no.~4, pp. 499--508, 2005.

\end{thebibliography}
\bibliographystyle{IEEEtran}

\end{document}